\documentclass[a4paper,conference]{IEEEtran}
\IEEEoverridecommandlockouts
\usepackage{geometry}
\usepackage{cite}
\usepackage{amsmath,amssymb,amsfonts}
\usepackage{graphicx}
\usepackage{textcomp}
\usepackage{xcolor}
\usepackage{algorithm}
\usepackage{algpseudocode}
\def\BibTeX{{\rm B\kern-.05em{\sc i\kern-.025em b}\kern-.08em
    T\kern-.1667em\lower.7ex\hbox{E}\kern-.125emX}}
 
\makeatletter
    \newcommand{\linebreakand}{%
      \end{@IEEEauthorhalign}
      \hfill\mbox{}\par
      \mbox{}\hfill\begin{@IEEEauthorhalign}
    }
    \makeatother

\begin{document}

\title{Automated Stitching of Coral Reef Images and Extraction of Features for Damselfish Shoaling Behavior Analysis
\thanks{R. R. Pineda is supported by a postgraduate scholarship from the Engineering Research and Development for Technology (ERDT), Philippines.}
}

\author{\IEEEauthorblockN{Riza Rae A. Pineda}
\IEEEauthorblockA{\textit{Department of Computer Science} \\
\textit{University of the Philippines}\\
Quezon City, Philippines \\
rapineda1@up.edu.ph}
\and
\IEEEauthorblockN{Kristofer E. delas Pe\~nas\thanks{K. delas Pe\~nas is supported by doctoral fellowships from the University of the Philippines and the Engineering Research and Development for Technology (ERDT), Philippines.}}
\IEEEauthorblockA{\textit{Department of Computer Science} \\
\textit{University of the Philippines}\\
Quezon City, Philippines \\
kedelaspenas@up.edu.ph}
\and
\IEEEauthorblockN{Dana P. Manogan}
\IEEEauthorblockA{\textit{Marine Science Institute} \\
\textit{University of the Philippines}\\
Quezon City, Philippines \\
dmanogan@msi.upd.edu.ph}

% \linebreakand

% \IEEEauthorblockN{Patrick C. Cabaitan}
% \IEEEauthorblockA{\textit{Marine Science Institute} \\
% \textit{University of the Philippines}\\
% Quezon City, Philippines \\
% pcabaitan@msi.upd.edu.ph}
% \and
% \IEEEauthorblockN{Patrick C. Cabaitan}
% \IEEEauthorblockA{\textit{Marine Science Institute} \\
% \textit{University of the Philippines}\\
% Quezon City, Philippines \\
% pcabaitan@msi.upd.edu.ph}
}

\maketitle

\begin{abstract}
Behavior analysis of animals involves the observation of intraspecific and interspecific interactions among various organisms in the environment. Collective behavior such as herding in farm animals, flocking of birds, and shoaling and schooling of fish provide information on its benefits on collective survival, fitness, reproductive patterns, group decision-making, and effects in animal epidemiology. In marine ethology, the investigation of behavioral patterns in schooling species can provide supplemental information in the planning and management of marine resources. Currently, damselfish species, although prevalent in tropical waters, have no adequate established base behavior information. This limits reef managers in efficiently planning for stress and disaster responses in protecting the reef. Visual marine data captured in the wild are scarce and prone to multiple scene variations, primarily caused by motion and changes in the natural environment. The gathered videos of damselfish by this research exhibit several scene distortions caused by erratic camera motions during acquisition. To effectively analyze shoaling behavior given the issues posed by capturing data in the wild, we propose a pre-processing system that utilizes color correction and image stitching techniques and extracts behavior features for manual analysis.
\end{abstract}


\section{Introduction}
Behavioral data of fish provide salient information in the analysis of fish maturity, biological fitness, reproductive and seasonal cycles, and food chains\cite{fisher}. This study may also provide appropriate behavioral models in the investigation of other vertebrates. Important fitness and survival benefits have been observed from the formation of mixed-species social groups among animals \cite{goodale}. Many species of fish participate in group living such as schooling or shoaling for increased predator avoidance, better foraging and reproductive opportunities, and optimized energy use \cite{abrahams,herskin,marras,weihs,snekser,nadler2018}. 

Damselfish species, which easily proliferate in reefs, have been observed to have detrimental effects to the recovery of coral reefs. This family of farmer fishes predominantly cultivate algal turfs and break juvenile corals to allow turf algae to grow\cite{schopmeyer}. However, collective information regarding how these species respond to their environment have yet to be completely established. Reef managers are currently unable to obtain adequate information for the effective planning of reef restoration, rehabilitation, and sustainability projects. 
% Gregarious damselfish species have been found to have reduced minimal metabolic rates when measured in a group than when measured alone. This is attributed to the increase in energy spent on vigilance and induced by the stress of isolation \cite{nadler2016}. 

At present, collecting behavior data in the wild is logistically and financially demanding, and physically exhausting. As events in the natural environment are inherently stochastic, sampling and capturing replicates are difficult and taxing. Currently, experts are only able to acquire snapshots of behaviors of the individuals since simulating a laboratory-like setup underwater is complex and often impractical.

Computer vision methods have been used to automate manual tasks in data preparation and processing. Corpuz et al. \cite{corpuz} proposed a rapid coral reef assessment system from towed camera arrays using mosaicking and an optimal seam search algorithms. With the capturing design of their method, translations and minimal rotations are expected from the collected images. Their method, as a result, may not perform well on data with high rotational variance.

Motivated by this technology and the need to establish more findings regarding damselfish species, we adapted the use of image correction and stitching techniques to provide more condesed details for damdeslfish behavior analysis. This research is in collaboration with an ongoing study on the shoaling behavior of damelfishes in coral reefs by the Marine Science Institute of the University of the Philippines. In the principal study \cite{dana}, we analyze shoaling patterns to understand the internal organization of coral reef shoaling species. We support this goal by providing a visual dataset pre-processing tool that primarily produces trajectory maps and computes for behavior features which may then be utilized for further analysis. 

Initially, we collected videos in two different reef sites in the Philippines. We then used an automatic white balancing algorithm to improve the color quality of the images. We labelled key points of objects and fish for each frame and submitted the frames for stitching, trajectory mapping, and behavior feature set extraction. Finally, we performed qualitative analysis and field expert validation on the resulting map and features. 
\section{Methodology}

\subsection{Dataset}                                    
\begin{figure}[htbp]
    \centering
    \includegraphics[width=0.48\linewidth]{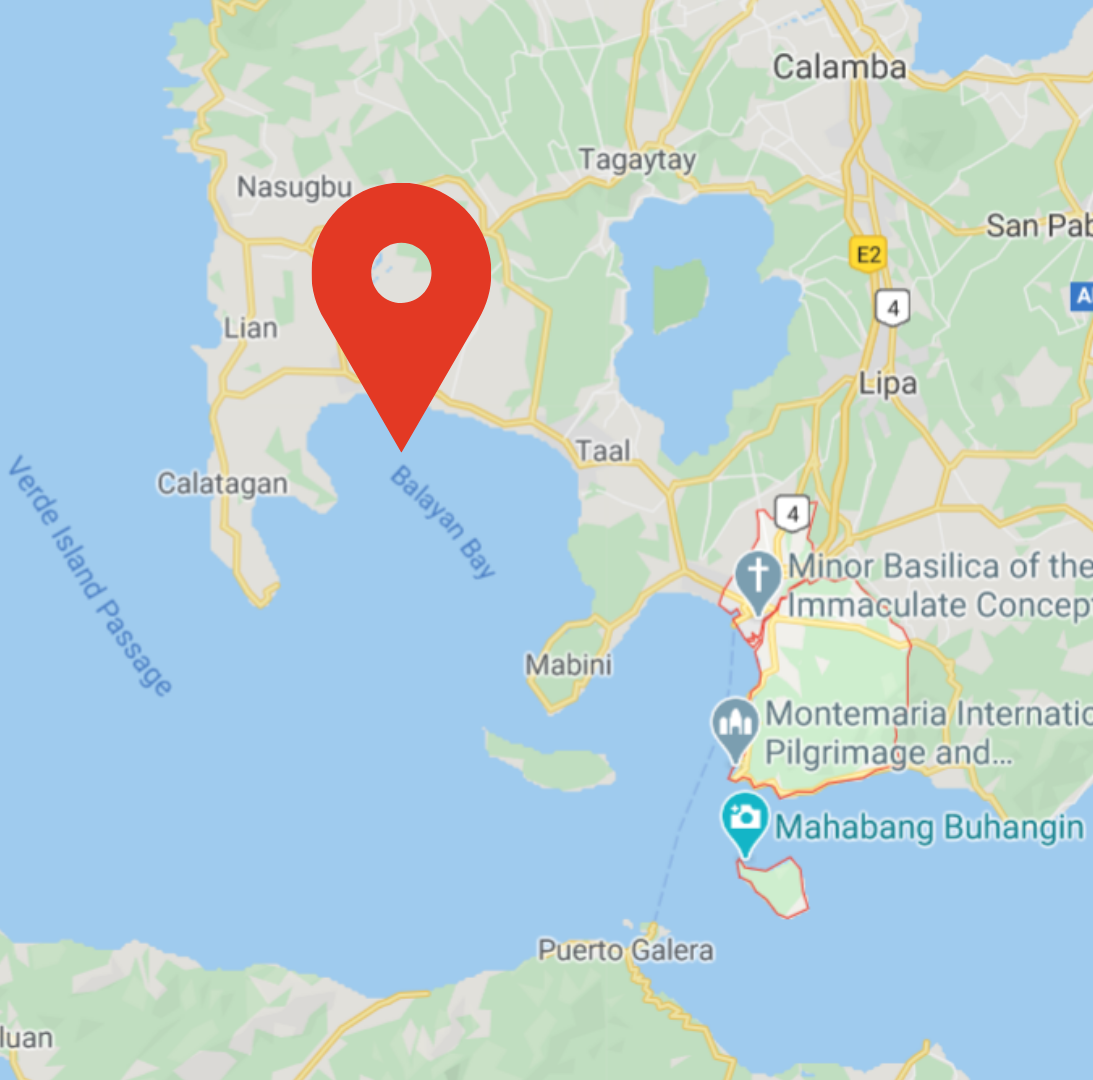}
    \includegraphics[width=0.48\linewidth]{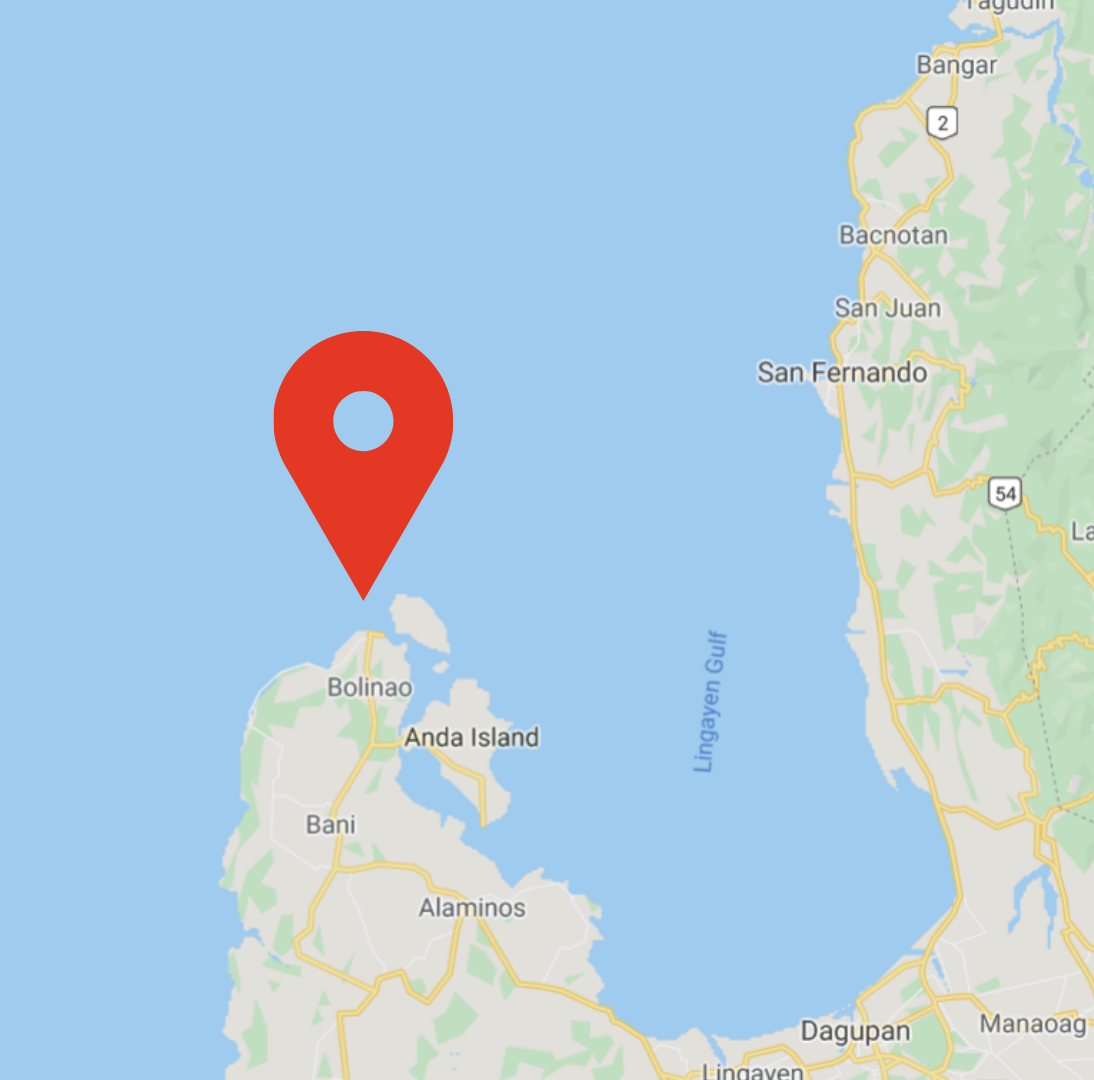}
    \vskip 0.12cm
    \includegraphics[width=0.48\linewidth]{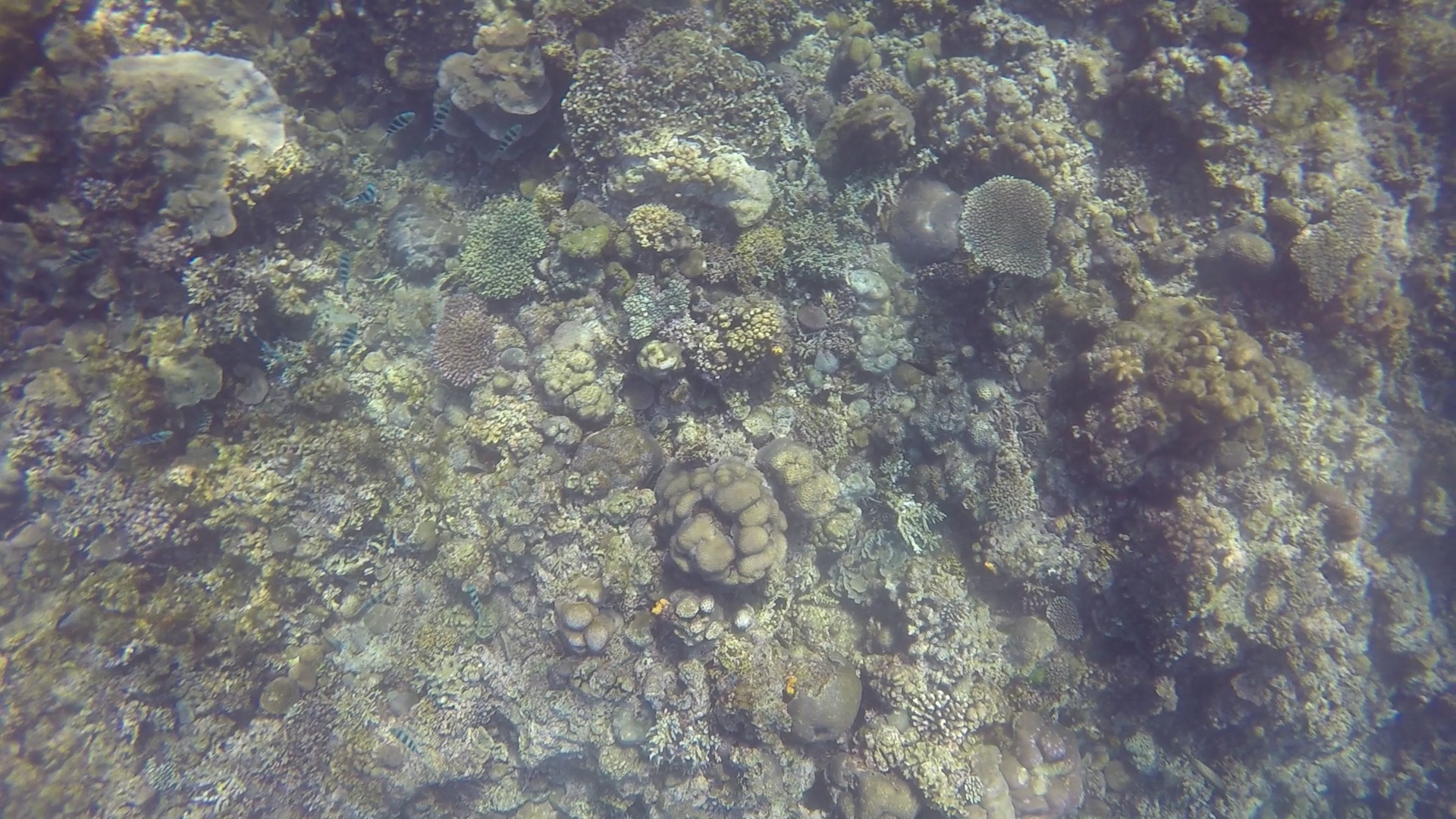}
    \includegraphics[width=0.48\linewidth]{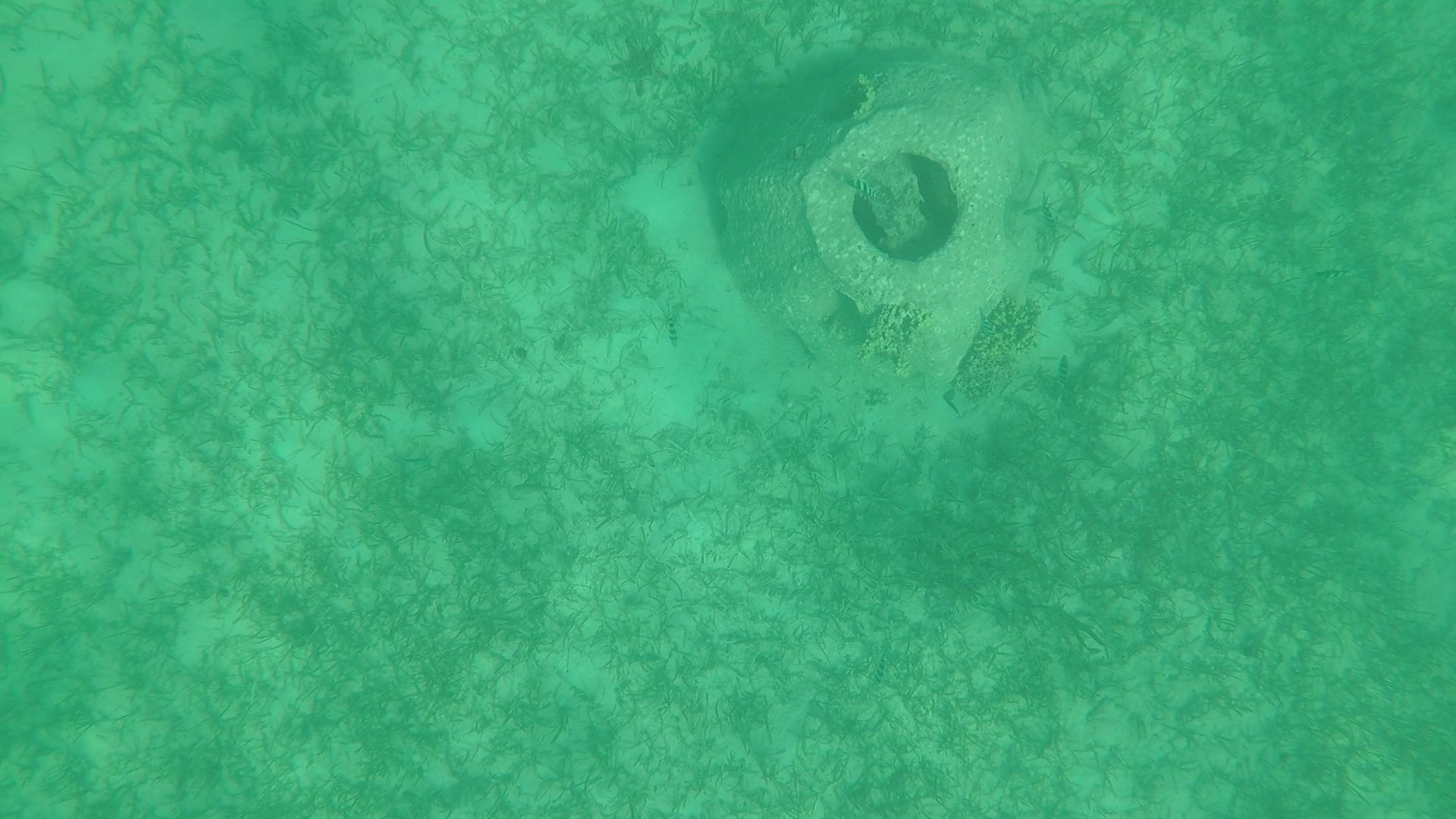}
    \caption[width=\linewidth]{Map locations (top row) of the two data sites, Batangas (left columun) and Bolinao (right column), and sample raw image collected in each site (bottom row) displaying the difference in scene conditions}
    \label{fig:bat_bol}
\end{figure}

A total of 25 videos from two underwater sites in the Philippines, Batangas and Bolinao, with significantly different conditions were collected. Batangas is known to be a hotspot for diving activities due to its marine resource-rich location. This province in the Philippines is part of the Verde Island Passage, which has been declared as the center or marine shorefish biodiversity in the world\cite{carpenter}. Bolinao, on the other hand, is an established location for fisheries. The specific site selected from this province is a predominantly sandy location. Shown in Fig. \ref{fig:bat_bol} is a visual comparison of the two selected sites. The videos were captured using a single handheld GoPro Hero 4 camera at approximately 3 to 4 meters from the target individuals.                   
%https://www.researchgate.net/publication/227112122_The_center_of_the_center_of_marine_shore_fish_biodiversity_The_Philippine_Islands
Key points for each frame with reference to the first frame of the video sequence were manually labelled. Fish points corresponding to the head, center, and tail were individually annotated post color correction. 
\subsection{Image Color Correction}
Frames were sequentially extracted from the videos at a rate of 3 frames per second. Due to varying environmental conditions during the actual collection of the videos, the images captured had differing illuminations and noise. We initially employed an automatic white balancing algorithm designed by Lam \cite{lam} for color correction prior to stitching. In Lam's work, he proposed a quadratic mapping of intensities based on the gray world assumption and the Retinex theory for automatic white balancing \cite{lam}.

\subsubsection{Gray World Assumption}
The underlying principle behind the gray world assumption is that the average of the negatives, which tends to be neutral, is biased towards dark regions of the image in film photography. For typical scenes, it is assumed that the average intensity of the red, green, and blue channels are equal \cite{lam}. This is given by,
\begin{equation}
    R_{avg} = \frac{1}{MN} \sum_{x=1}^{M}\sum_{y=1}^{N}I_r(x,y)
\end{equation}
\begin{equation}
    G_{avg} = \frac{1}{MN} \sum_{x=1}^{M}\sum_{y=1}^{N}I_b(x,y)
\end{equation}
\begin{equation}
    B_{avg} = \frac{1}{MN} \sum_{x=1}^{M}\sum_{y=1}^{N}I_b(x,y)
\end{equation}

where I is the \(M \times N\) image, \(r\) is the red channel, \(b\) is the blue channel, \(g\) is the green channel, and \((x,y)\) are the 2D coordinates of each pixel.

Solely applying averaging does not generally produce good results. Therefore, the green channel is usually ignored and gain values \(\alpha\) and \(\beta\) of the red and blue channels, respectively, are computed for color correction \cite{lam}. The gain values are computed as,

\begin{equation}
    \alpha = \frac{G_{avg}}{R_{avg}}
\end{equation}
\begin{equation}
    \beta = \frac{G_{avg}}{B_{avg}}
\end{equation}
We then apply \(\alpha\) and \(\beta\) on the original image \(I\) for correction using the equations below:

\begin{equation}
    \hat{I_r}(x,y) = \alpha I_r(x,y)
\end{equation}
\begin{equation}
    \hat{I_b}(x,y) = \beta I_b(x,y)
\end{equation}.

\subsubsection{Retinex Theory}
The Retinex theory states that the perceived white of humans is associated with the maximum cone signals of the eyes. To perform white balancing on images, the maximum values of the three color channels red, green, and blue must then be equalized. Similar to the gray world assumption, we use the red and blue channels and adjust the pixel values by multiplying the gain values to their respective channel \cite{lam}.

\subsubsection{Automatic White Balance}
\begin{figure}[h]
    \centering
    \includegraphics[width=0.48\linewidth]{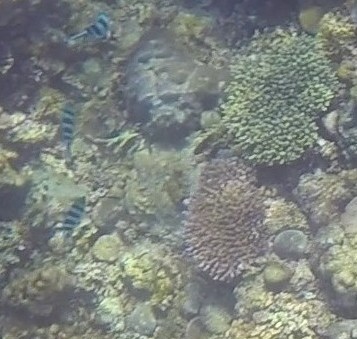}
    \includegraphics[width=0.48\linewidth]{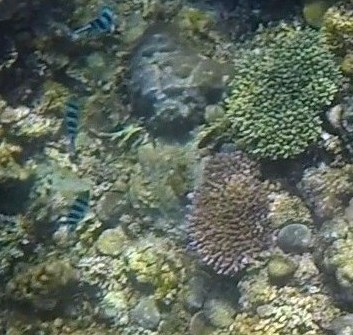}
    \vskip 0.12cm
   \includegraphics[width=0.48\linewidth]{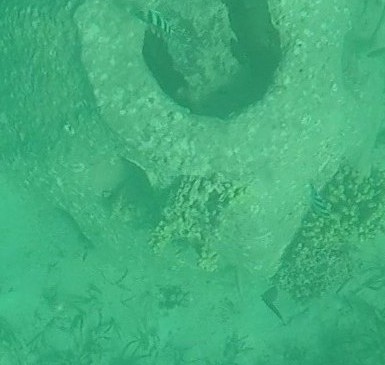}
    \includegraphics[width=0.48\linewidth]{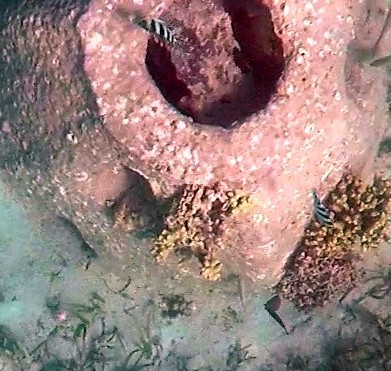}
    \caption{A more detailed look on the effect of performing automatic white balancing on the collected images (raw images on the left; corrected images on the right).
     We see more recognizable objects richer in color as a consequence to applying \cite{lam}.}
    \label{fig:colorcorr}
\end{figure}

Combining the the gray world assumption and the retinex theory, the automatic white balance parameters \(\mu\) and \(v\) are given by:

\begin{equation}
\begin{bmatrix}
\sum \sum I_{r}^{2} & \sum \sum I_{r}\\
max I_{r}^{2} & max I_{r}
\end{bmatrix} \begin{bmatrix}
\mu_r\\
v_r
\end{bmatrix} = \begin{bmatrix}
\sum \sum I_{g}\\
max I_{g}
\end{bmatrix}
\end{equation}

\begin{equation}
\begin{bmatrix}
\sum \sum I_{b}^{2} & \sum \sum I_{b}\\
max I_{b}^{2} & max I_{b}
\end{bmatrix} \begin{bmatrix}
\mu_b\\
v_b
\end{bmatrix} = \begin{bmatrix}
\sum \sum I_{g}\\
max I_{g}
\end{bmatrix}
\end{equation}
where \(r\) and \(b\) pertain to the red and blue channel, respectively. Forming this as a system of linear equations with two unknowns, \(\mu\) and \(v\) for the two color channels can be solved by using Gaussian elimination \cite{lam}.
Fig. \ref{fig:colorcorr} shows the original and color-corrected versions of a sample from each collected site. We also show the histogram of each image in Fig. \ref{fig:colorcorr} to provide more information on the differences in tonal distributions between the raw and corrected images. 

The resulting images were then passed to the stitching module of the system to generate a reef map. 
\begin{figure*}[h]
    \centering
    \includegraphics[width=0.4\linewidth]{figures/Bat_orig.jpg}
    \includegraphics[width=0.4\linewidth]{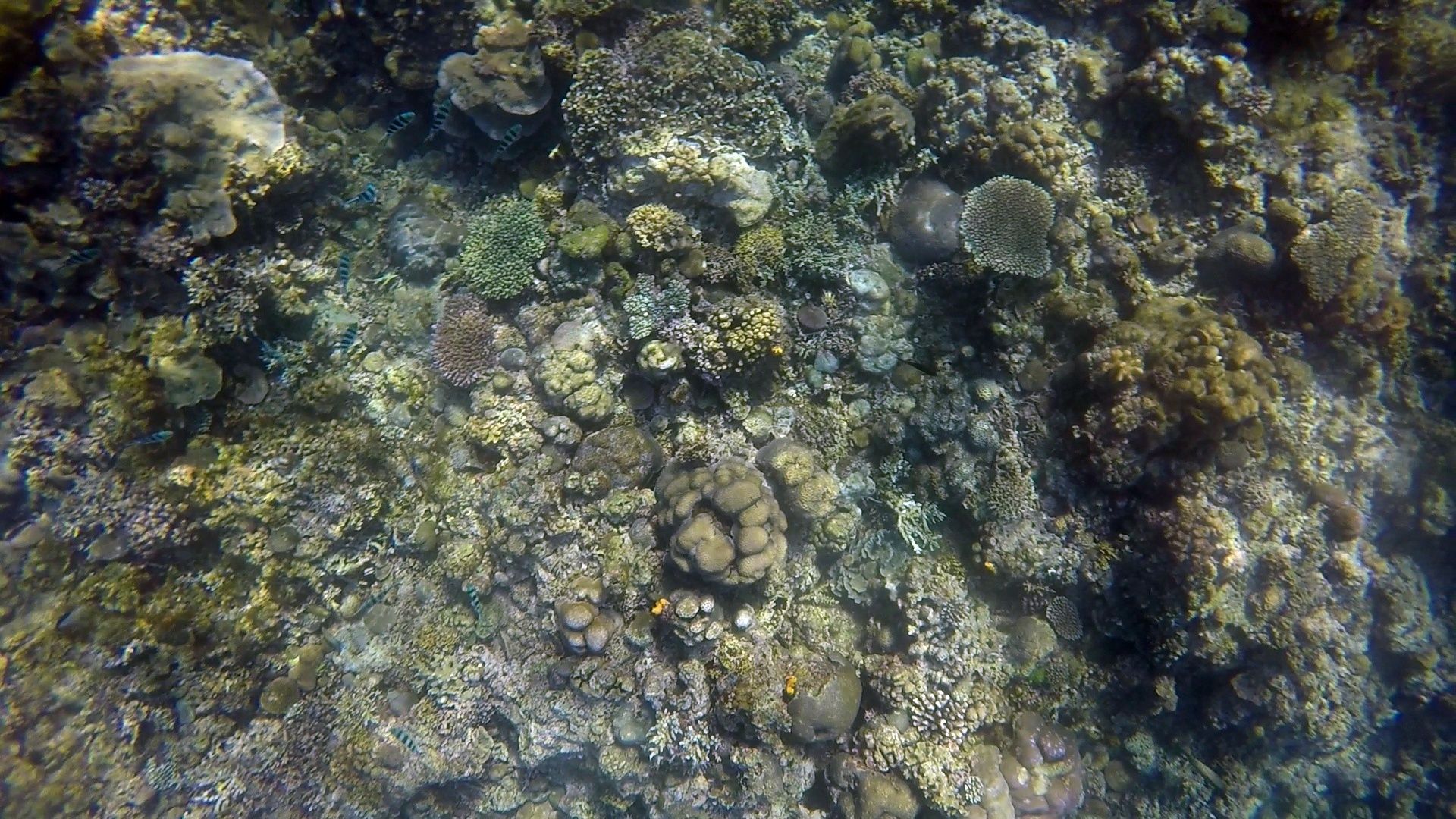}
    \vskip 0.12cm
    \includegraphics[width=0.4\linewidth]{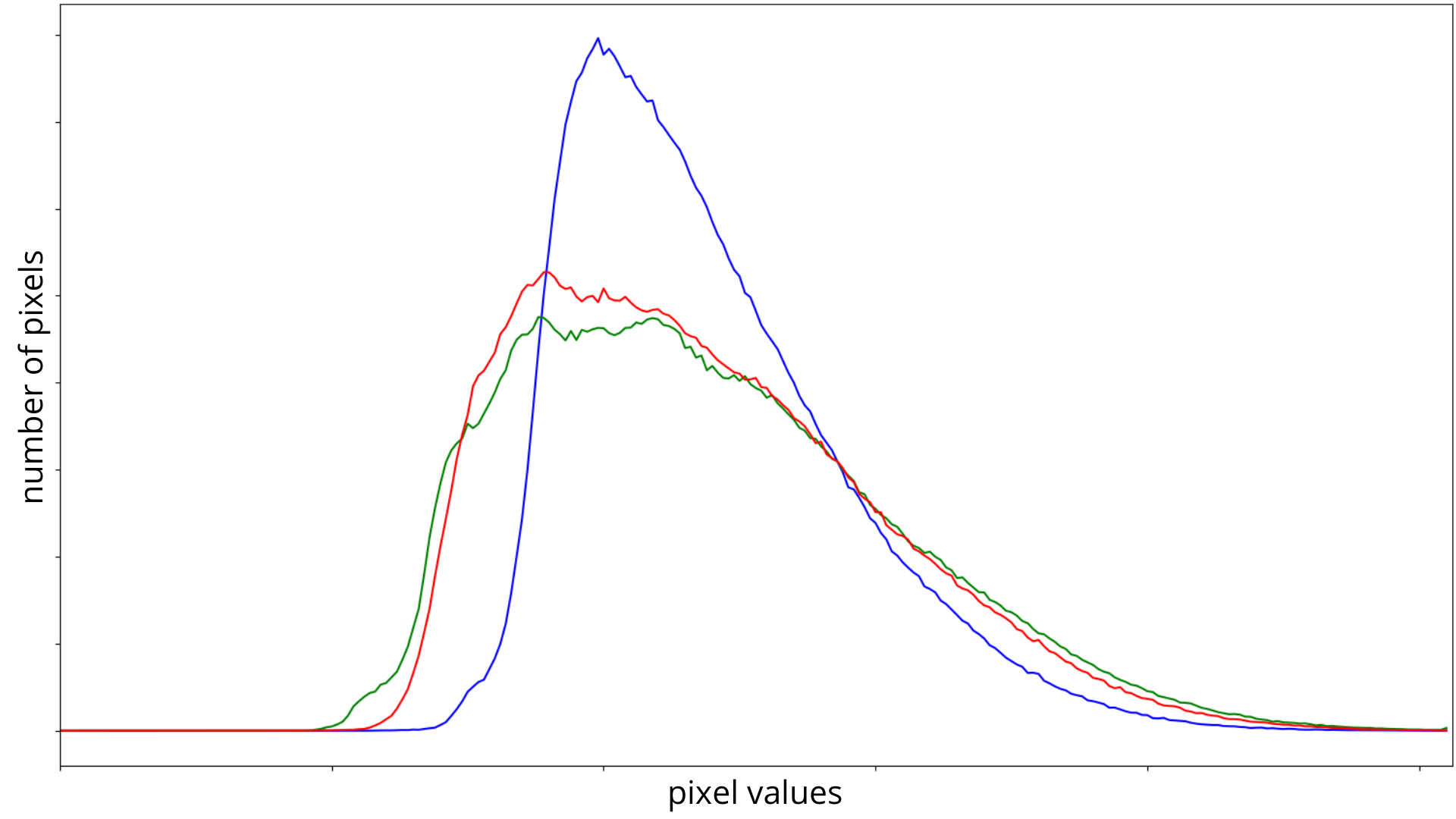}
    \includegraphics[width=0.4\linewidth]{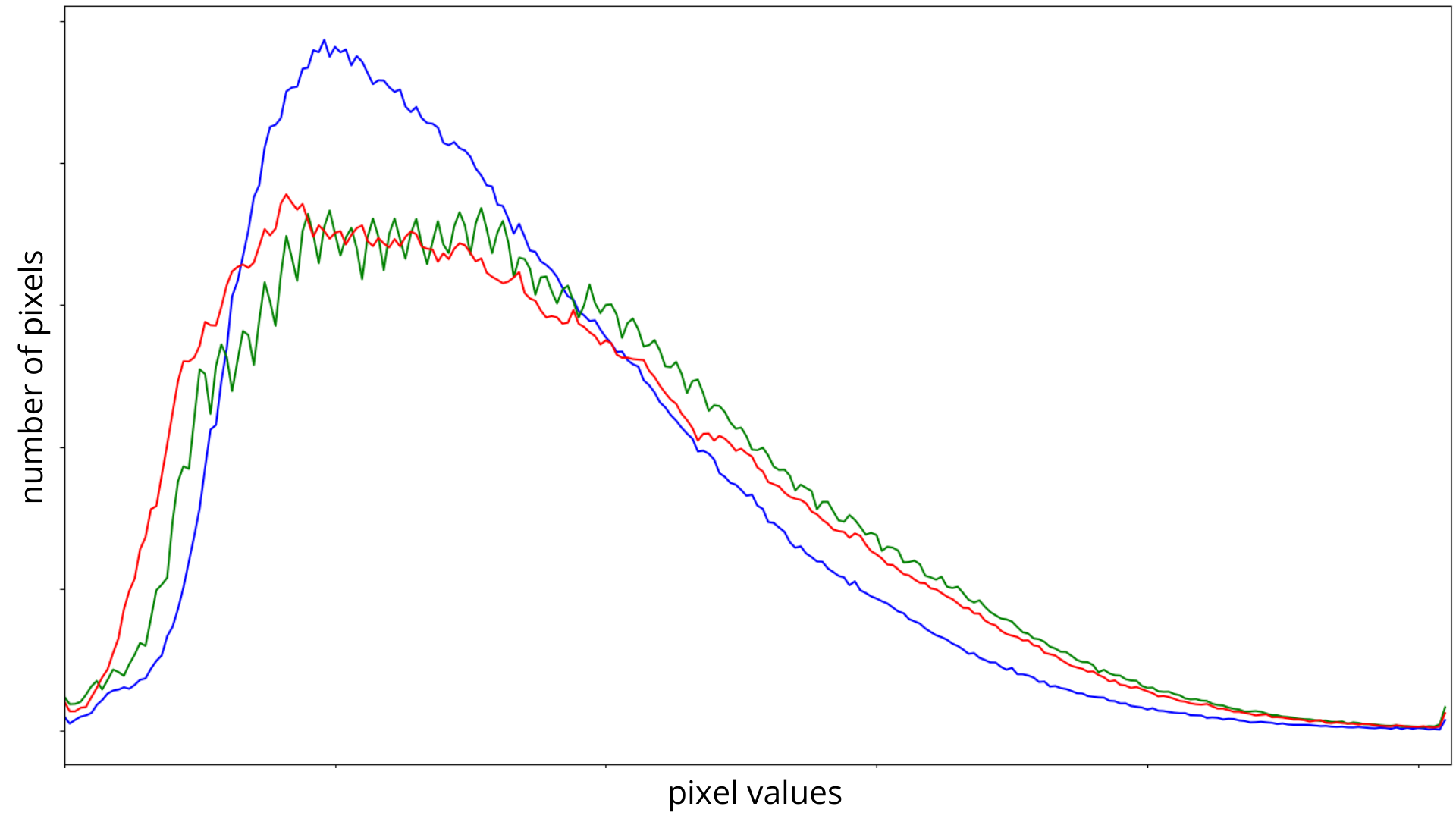}
    \vskip 0.12cm
   \includegraphics[width=0.4\linewidth]{figures/Bol_orig.jpg}
    \includegraphics[width=0.4\linewidth]{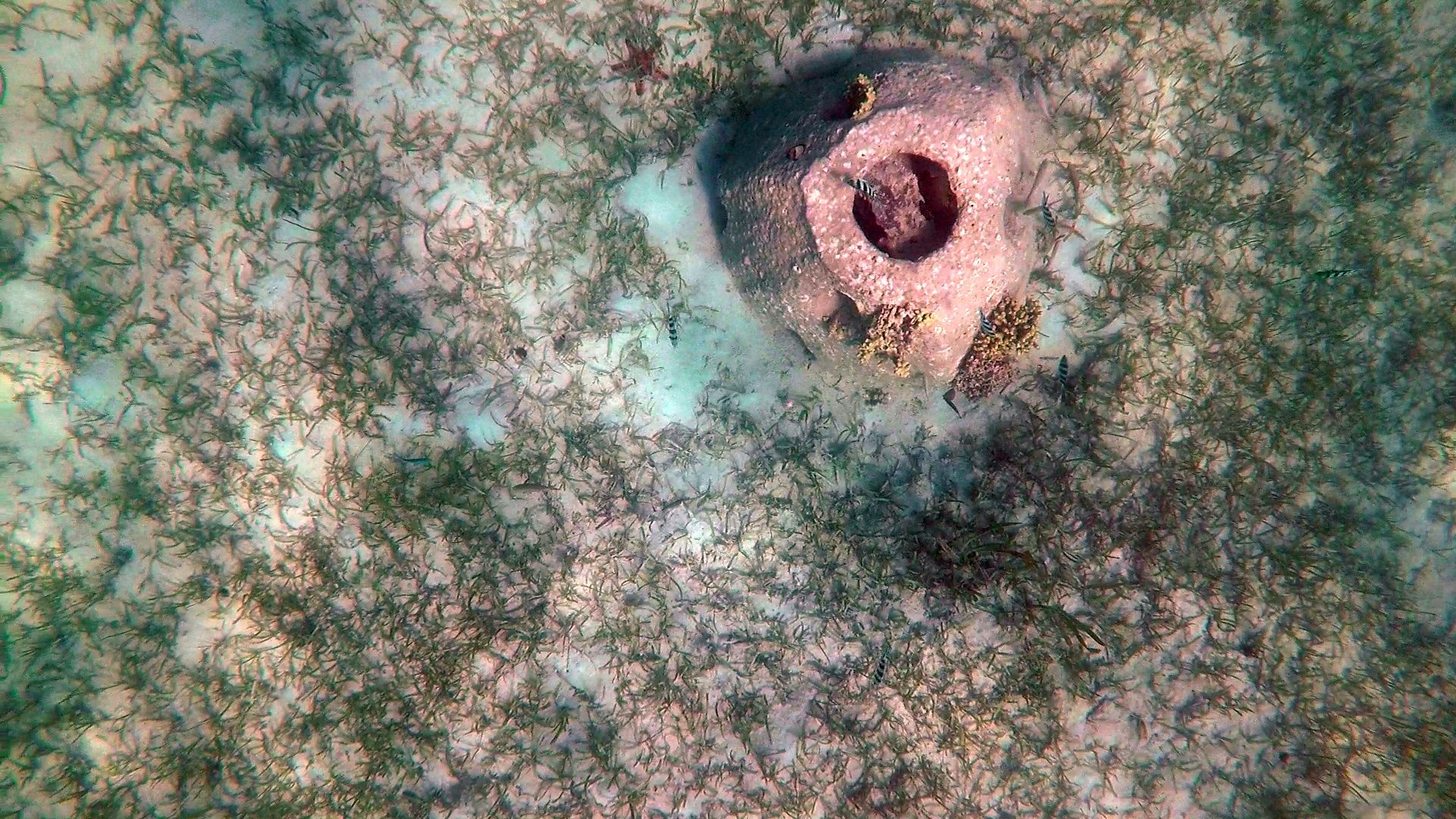}
     \vskip 0.12cm
   \includegraphics[width=0.4\linewidth]{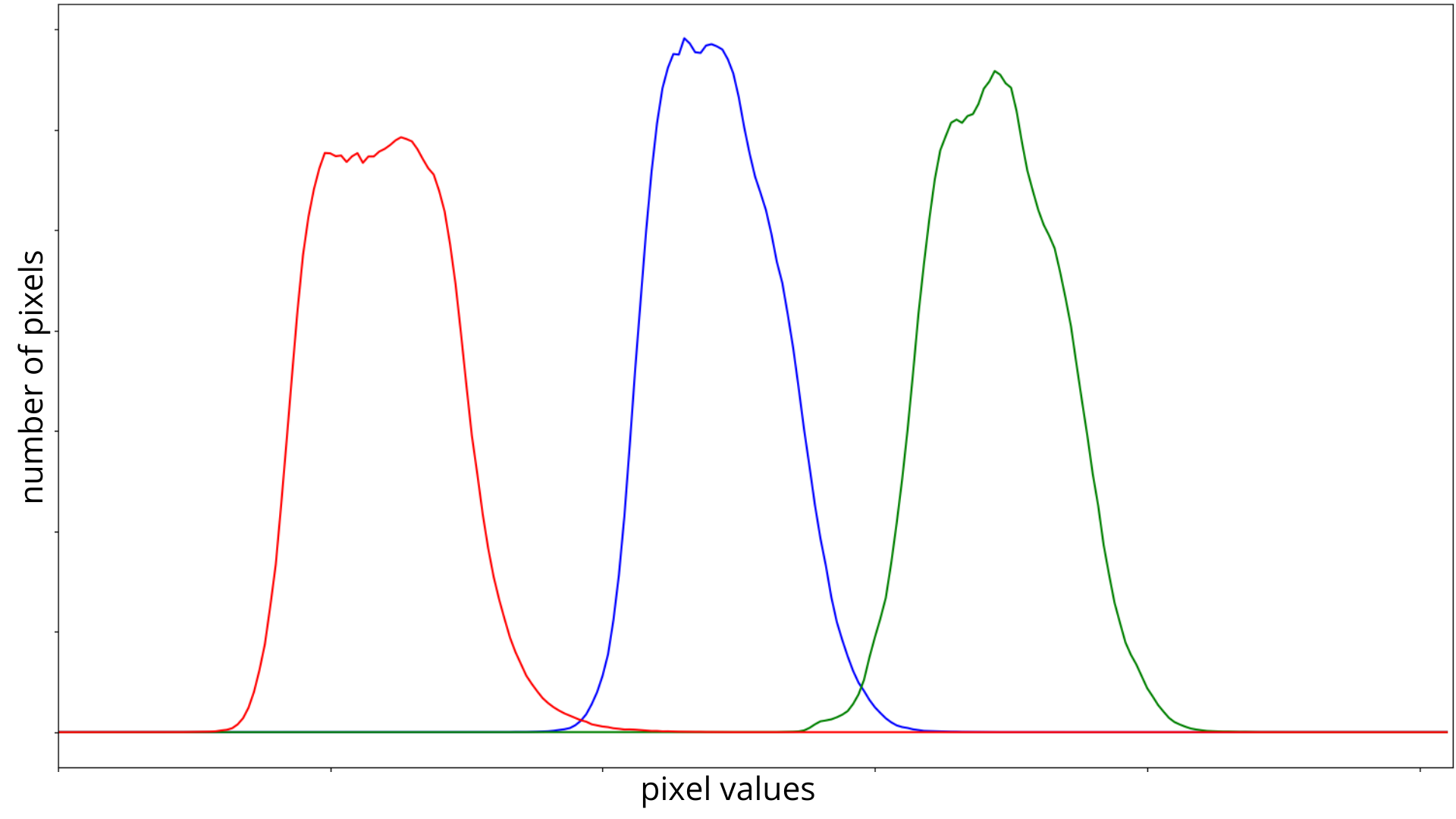}
    \includegraphics[width=0.4\linewidth]{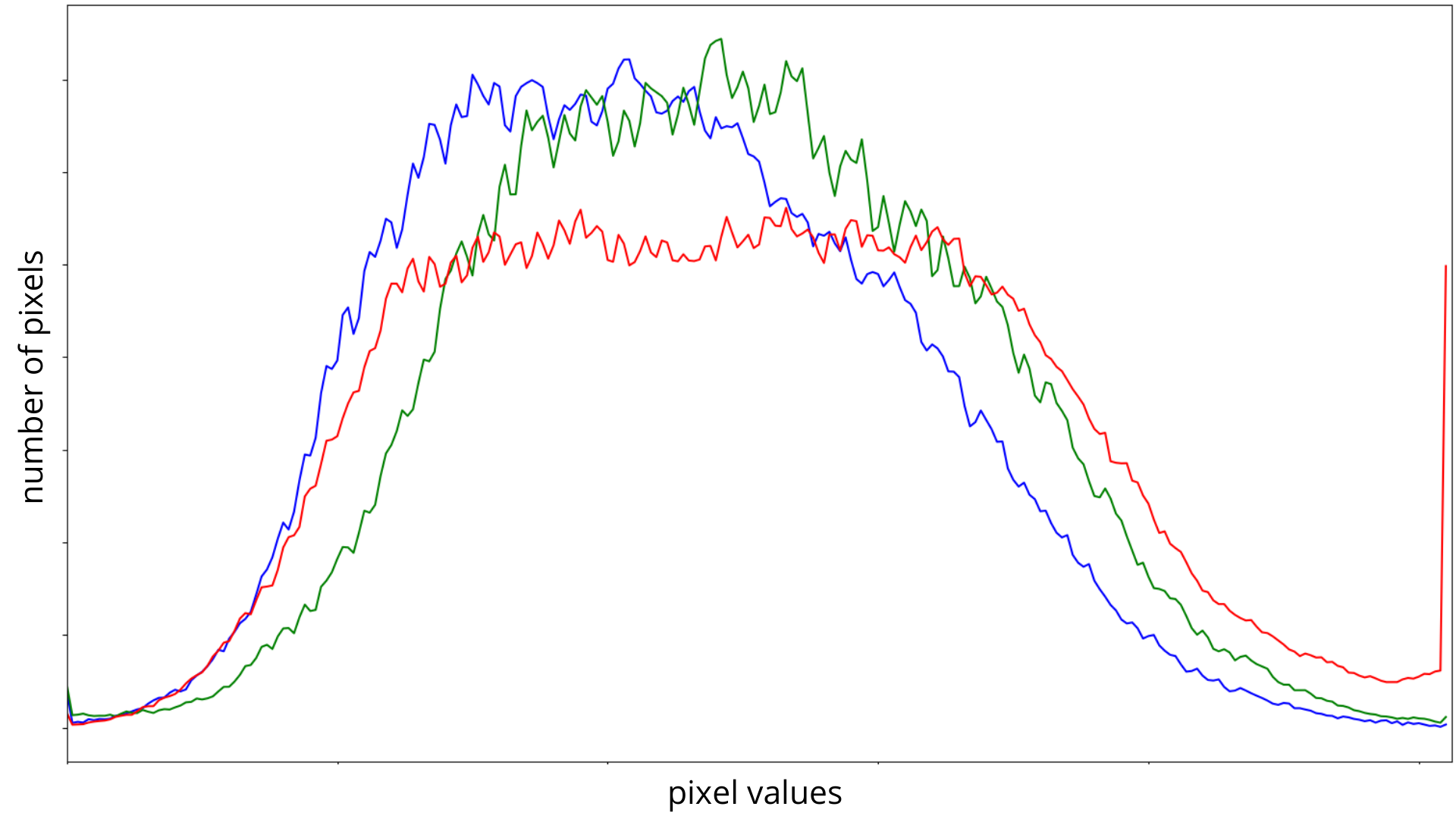}
    \caption{Effect of applying the automatic white balancing algorithm of \cite{lam} in samples collected from the two data sites. The left column pertains to the raw information while the figures on the right column show the corrected information. For a densely noisy image, as in the lower left images, we observe a large disparity among the different channels. We perceive noticeable improvements in color after applying \cite{lam}.}
    \label{fig:dataset}
\end{figure*}
 
\subsection{Image Stitching}
We performed image stitching through the steps defined in Algorithm \ref{alg:stitching}. Once the sequence of frames were presented to the stitching function, transformation matrices were automatically computed using the Random Sampling Consensus (RANSAC) algorithm \cite{fischler} for each succeeding frame relative to the initial frame. 
\begin{algorithm}
\caption{Image Stitching}
\begin{algorithmic}[1]
    \Procedure{Image Stitching}{}
    \State Estimate the affine transformation matrix \(M\) of the current frame relative to the initial frame using RANSAC
    \State Compute for the dimensions of the stitched map
    \State Pad the images and their corresponding transformation matrices
    \State Warp the current frame using the shifted transformation matrix
    \State Combine the warped images and the initial frame
    \State Apply morphological closing to remove black pixels between the overlay regions of succeeding frames
\EndProcedure
\end{algorithmic}
\label{alg:stitching}
\end{algorithm}
% \begin{enumerate}
%     \item Estimation of the affine transformation matrix \(M\) of the current frame relative to the initial frame using Random Sampling Consensus (RANSAC) \cite{fischler}
%     \item Computation of the dimensions of the stitched map
%     \item Padding of the images and their corresponding transformation matrices
%     \item Affine warping of the current frame using the shifted transformation matrix
%     \item Combination of the warped images and the initial frame
%     \item Application of morphological closing to remove black pixels between the overlay regions of succeeding frames
    
% \end{enumerate}

The RANSAC \cite{fischler} algorithm is a predictive model that generates candidate solutions using the minimum number of observations needed to estimate the underlying model parameters\cite{fischler, derpanis_2010}. This algorithm is described in Algorithm \ref{alg:ransac}.

\begin{algorithm}
\caption{Random Sampling Consensus (RANSAC)\cite{fischler,derpanis_2010}}
\begin{algorithmic}[1]
    \Procedure{RANSAC}{}
    \State Randomly select the minimum number of points required to determine the model
    parameters
    \State Solve for the parameters of the model
    \State Determine how many points from the set of all points fit with a predefined tolerance \(\epsilon\)
    \State If the fraction of the number of inliers over the total number points in the set
    exceeds a predefined threshold \(\tau\), re-estimate the model parameters using all the
    identified inliers and terminate.
    \State Otherwise, repeat steps 1 through 4 (maximum of N times)
\EndProcedure
\end{algorithmic}
\label{alg:ransac}
\end{algorithm}

\begin{figure}[h]
    \centering
    \includegraphics[width=0.98\linewidth]{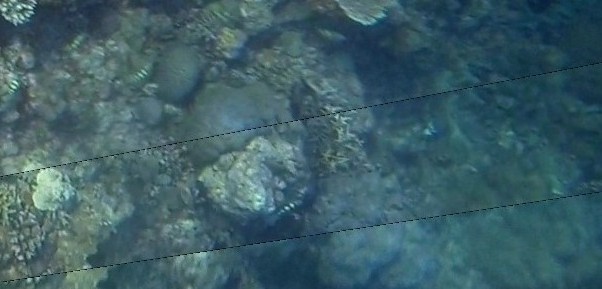}
    \vskip 0.12cm
    \includegraphics[width=0.98\linewidth]{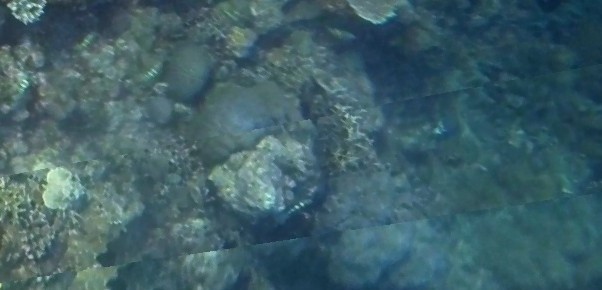}
    \caption{Result of applying morphological closing on the stitched images. The top image contains black pixels as a result of warping and padding. We remove this by applying dilation and erosion on the whole image (bottom).}
    \label{fig:morph}
\end{figure}

After the computation of the affine transformation matrices, the maximum and minimum dimensions from all the transformations were computed to determine the padding values. The initial frame was then padded and the transformation matrices of the succeeding images were shifted using the same padding parameters. Iteratively, each shifted affine transformation matrix was applied to its corresponding frame and superimposed on the current partial stitched map. As a result, we obtained an overall stitched map and smoothened the edges between the images by applying morphological closing. This method is a two-step procedure where the image is initially dilated and then eroded. Fig. \ref{fig:morph} shows the difference between performing morphological closing on a subset of a sample stitched map. 

The resulting image was then used to plot fish trajectories. Since fish points were manually labelled from the original orientation of each frame, a new set of corresponding points must be calculated for the stitched map. We performed this adjustment by applying the dot product of the shifted transformation matrix of the current frame to the original set points. This affine transformation method is given by
\begin{equation}
    [x_{new},y_{new}] = M_{shifted} \cdot [x_{orig},y_{orig},1]^{T}
\end{equation}
where \(x_{new}\) and \(y_{new}\) are the new fish coordinates in the transformed image, \(M_{shifted}\) is the translated \(2\times3\) transformation matrix of the current image, and \(x_{orig}\) and \(y_{orig}\) are the original fish coordinates.

\begin{figure*}[h]
    \centering
    \includegraphics[width=0.5\linewidth]{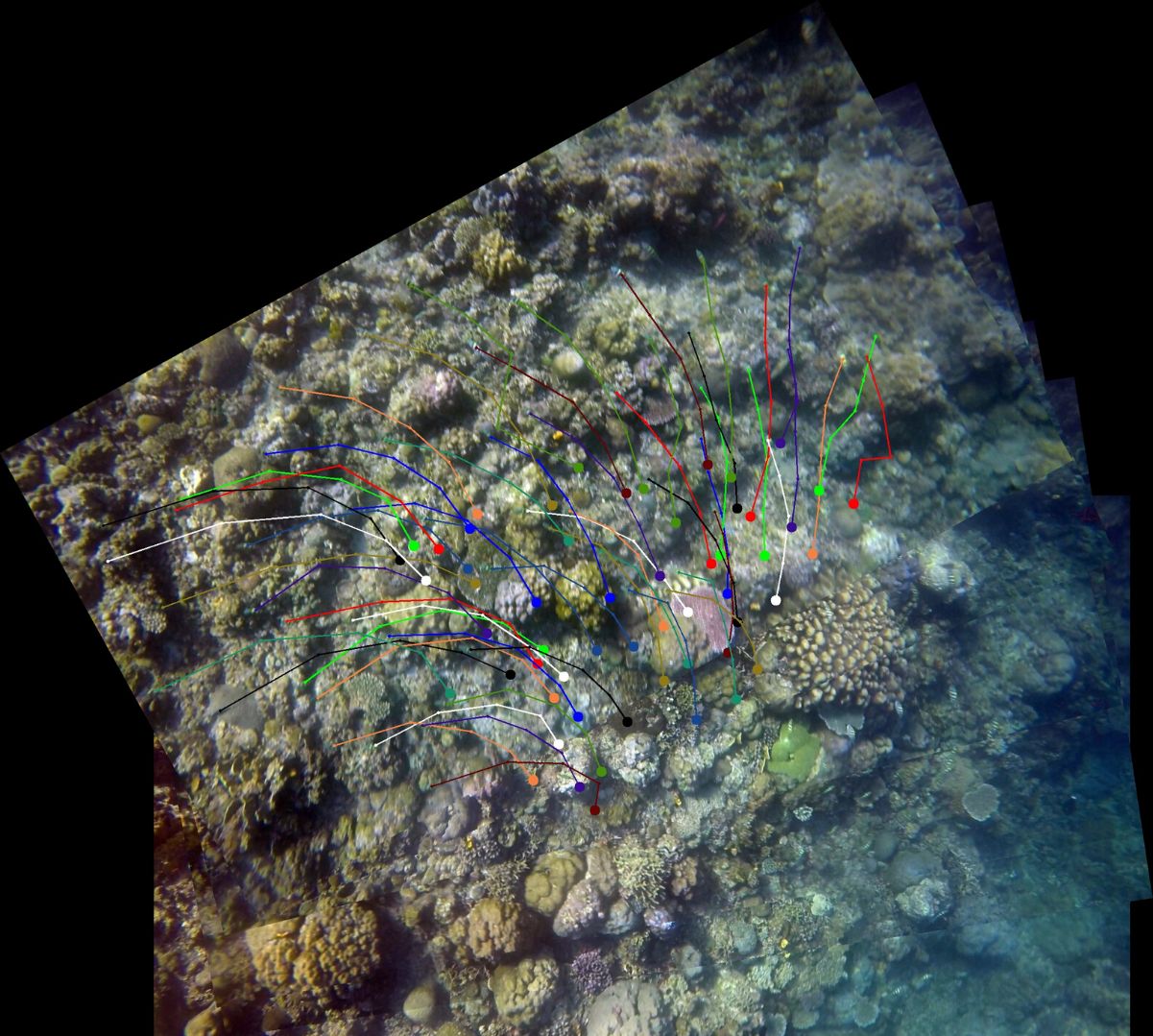}
    \includegraphics[width=0.475\linewidth]{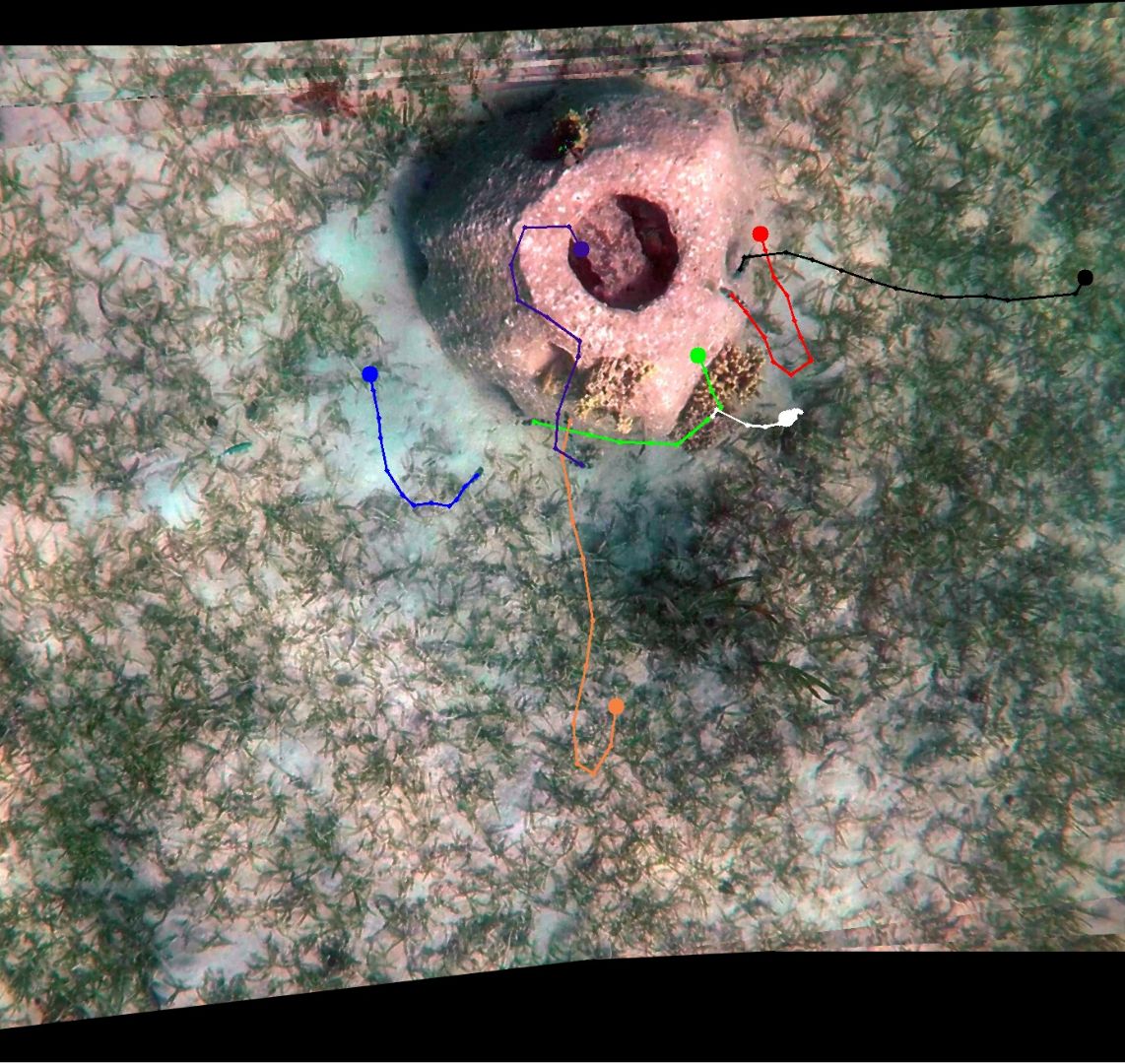}
    \caption{Results of the stitching function. The top and bottom images are stitched maps generated from videos collected from the Batangas and Bolinao sites, respectively.}
    \label{fig:mapwithtraj}
\end{figure*}

 The overall result of this section is a stitched map with plotted trajectories for each fish, shown in Fig. \ref{fig:mapwithtraj}.

\subsection{Shoaling Behavior Feature Extraction}
After generating a trajectory segment for each individual in the video, we proceeded to the extraction of behavior features. The identified features are spatial distance travelled, overall displacement, average speed, estimated body length of each fish, and the distance, angle, and species of the head of each fish to the nearest fish. 
\subsubsection{Total distance travelled}
The total distance travelled is given by the sum of all the Euclidean distances of each segment in the trajectory. This feature is measured in pixels.
\subsubsection{Overall displacement}
The overall displacement is the beeline distance from the start point to the end point.
\subsubsection{Average Speed}
The average speed, measured in pixels per second, is computed by dividing the total distance travelled by the total time travelled in seconds.
\subsubsection{Body Length}
Given three labelled points for each fish, head, center, and tail, the body length is computed as the sum of the lengths from the head to the center and the center to the tail. This feature is measured in pixels. 
\subsubsection{Nearest Neighbor-based Features}
\begin{figure}[htbp]
\centering
    \includegraphics[width=0.5\linewidth]{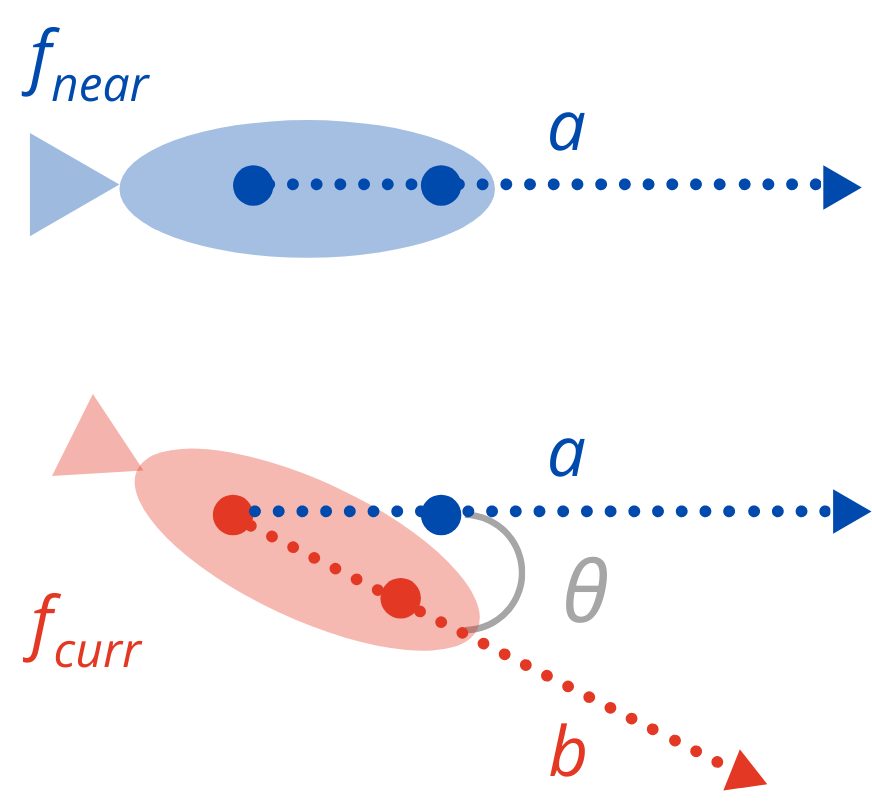}
\caption{Diagram for the nearest neighbor-based features. The blue circles correspond to the center and head points of \(f_{near}\). \(a\) is the heading direction of \(f_{near}\). \(b\) and the red dots are the heading direction, center point, and head point of \(f_{curr}\), respectively.}
\label{fig:rieucau}
\end{figure}

Rieucau et al. \cite{rieucau} identified two major features in the analysis of shoaling behavior in the wild. For each individual, he extracted the distance \(d\) to the nearest fish and the heading angle \(\theta\) with respect to the heading direction of the nearest fish.
For each individual \(f_{curr}\) at each time frame, the distance \(d\) to the nearest fish \(f_{near}\) is the Euclidean distance and the heading angle \(\theta\) is computed with respect to its heading direction \(b\) and the heading direction of the nearest fish \(a\) using the inverse cosine function. 
Fig. \ref{fig:rieucau} is a visualization of the computed parameters\cite{rieucau} for nearest neighbor-based features.

\section{Results and Discussion}\label{AA}
\subsubsection{Automatic White Balancing}
Shown in Figures \ref{fig:dataset} and \ref{fig:colorcorr} are the significant improvements in the color quality of the images after performing automatic white balancing. The resulting images provide more contrasting information on the color of the damselfish individuals. Objects such as corals and grasses visually appear closer to their actual color than the raw data. In Fig. \ref{fig:dataset}, the application of color correction significantly improved the color of both sample images. In the raw Bolinao videos, we can see frames with almost unrecognizable objects in the scene, especially in turbid waters. Inspecting further by looking into their histogram, we observe large disparities among the different color channels in these frames. Applying the method in \cite{lam}, the disparity in the color distribution is reduced, resulting in frames with higher contrast. In these color-corrected frames, the color profile of the sea grasses are comparable to that of images in non-turbid waters.

\subsubsection{Image Stitching}
The stitched maps appear cohesive and near-seamless with negligible incorrect patches. As shown in Fig. \ref{fig:mapwithtraj}, the trajectories displayed in the maps appear smooth and consistent despite multiple camera jitters and shaking during capture. This system is able to produce these results despite having only a single handheld camera as its capturing device. This makes this tool a viable alternative to more complex systems, such as those using rigs and stereo cameras, especially in instances where the travel load capacity for field works is limited.

The features generated appeared consistent with the visual information in the images and have been validated by our field expert. 

The end-to-end automated process described in this research took only miliseconds in execution per video with 15 significant frames. This speed in processing videos demonstrates the potential of this work in high-throughput generation of shoaling behavioral features from marine data for further analysis of marine scientists. 
\section{Conclusion}
We have shown in this paper a practical tool for pre-processing erratic visual data for marine scientists. Using our developed system, we remove the necessity of using specialized equipment other than a single underwater camera. This enhances the practicality and portability of the software most especially in conducting remote field work. This system hastens the data processing pipeline of marine scientists by producing summarized data without the need for manual evaluation of each video.

\section{Future Work}\label{SCM}
This research is in its early stages and multiple developments may be introduced to improve the tool. Full end to end automation can be introduced by improving the key point detection algorithm for stitching and using an automatic detector of fish.

\section{Authors}
R. R. A. Pineda is currently a doctoral student majoring in information science and engineering in Nara Institute of Science and Technology, Japan. Her research interests lie in computer vision and animal behavior studies.

K. E. delas Penas is currently a doctoral student in the University of Oxford, United Kingdom, working on computer vision and biomedical image analysis. 

D. P. Manogan is a graduating master's student from the Marine Science Institute of the University of the Philippines, Philippines. Her focus area is fish biology. She has extensive experience in conducting fish surveys and species identification.

% P. C. Cabaitan, Ph.D is an Assistant Professor at the Marine Science Institute of the University of the Philippines. His research interests are community ecology, restoration ecology, and giant clam conservation.
\bibliographystyle{IEEEtran}
\bibliography{ref}

\begin{thebibliography}{10}
\providecommand{\url}[1]{#1}
\csname url@samestyle\endcsname
\providecommand{\newblock}{\relax}
\providecommand{\bibinfo}[2]{#2}
\providecommand{\BIBentrySTDinterwordspacing}{\spaceskip=0pt\relax}
\providecommand{\BIBentryALTinterwordstretchfactor}{4}
\providecommand{\BIBentryALTinterwordspacing}{\spaceskip=\fontdimen2\font plus
\BIBentryALTinterwordstretchfactor\fontdimen3\font minus
  \fontdimen4\font\relax}
\providecommand{\BIBforeignlanguage}[2]{{%
\expandafter\ifx\csname l@#1\endcsname\relax
\typeout{** WARNING: IEEEtran.bst: No hyphenation pattern has been}%
\typeout{** loaded for the language `#1'. Using the pattern for}%
\typeout{** the default language instead.}%
\else
\language=\csname l@#1\endcsname
\fi
#2}}
\providecommand{\BIBdecl}{\relax}
\BIBdecl

\bibitem{fisher}
R.~Fisher, Y.-H. Chen-Burger, D.~Giordano, L.~Hardman, and F.-P. Lin,
  \emph{\BIBforeignlanguage{English}{Fish4Knowledge: Collecting and Analyzing
  Massive Coral Reef Fish Video Data}}, ser. Intelligent Systems Reference
  Library.\hskip 1em plus 0.5em minus 0.4em\relax Springer International
  Publishing, 2016.

\bibitem{goodale}
\BIBentryALTinterwordspacing
E.~Goodale, P.~Ding, X.~Liu, A.~Mart{\'i}nez, X.~Si, M.~Walters, and S.~K.
  Robinson, ``The structure of mixed-species bird flocks, and their response to
  anthropogenic disturbance, with special reference to east asia,'' \emph{Avian
  Research}, vol.~6, no.~1, p.~14, Aug 2015. [Online]. Available:
  \url{https://doi.org/10.1186/s40657-015-0023-0}
\BIBentrySTDinterwordspacing

\bibitem{abrahams}
M.~Abrahams and P.~Colgan, ``Risk of predation, hydrodynamic efficiency and
  their influence on school structure,'' \emph{Environmental Biology of
  Fishes}, vol.~13, pp. 195--202, 07 1985.

\bibitem{herskin}
\BIBentryALTinterwordspacing
J.~Herskin and J.~F. Steffensen, ``Energy savings in sea bass swimming in a
  school: measurements of tail beat frequency and oxygen consumption at
  different swimming speeds,'' \emph{Journal of Fish Biology}, vol.~53, no.~2,
  pp. 366--376, 1998. [Online]. Available:
  \url{https://onlinelibrary.wiley.com/doi/abs/10.1111/j.1095-8649.1998.tb00986.x}
\BIBentrySTDinterwordspacing

\bibitem{marras}
\BIBentryALTinterwordspacing
S.~Marras, S.~S. Killen, J.~Lindstr{\"o}m, D.~J. McKenzie, J.~F. Steffensen,
  and P.~Domenici, ``Fish swimming in schools save energy regardless of their
  spatial position,'' \emph{Behavioral Ecology and Sociobiology}, vol.~69,
  no.~2, pp. 219--226, Feb 2015. [Online]. Available:
  \url{https://doi.org/10.1007/s00265-014-1834-4}
\BIBentrySTDinterwordspacing

\bibitem{weihs}
\BIBentryALTinterwordspacing
D.~WEIHS, ``Hydromechanics of fish schooling,'' \emph{Nature}, vol. 241, no.
  5387, pp. 290--291, Jan 1973. [Online]. Available:
  \url{https://doi.org/10.1038/241290a0}
\BIBentrySTDinterwordspacing

\bibitem{snekser}
J.~Snekser, N.~Ruhl, K.~Bauer, and S.~Mcrobert, ``The influence of sex and
  phenotype on shoaling decisions in zebrafish,'' \emph{International journal
  of comparative psychology / ISCP; sponsored by the International Society for
  Comparative Psychology and the University of Calabria}, vol.~23, pp. 70--81,
  01 2010.

\bibitem{nadler2018}
\BIBentryALTinterwordspacing
L.~E. Nadler, S.~S. Killen, P.~Domenici, and M.~I. McCormick, ``Role of water
  flow regime in the swimming behaviour and escape performance of a schooling
  fish,'' \emph{Biology Open}, vol.~7, no.~10, 2018. [Online]. Available:
  \url{https://bio.biologists.org/content/7/10/bio031997}
\BIBentrySTDinterwordspacing

\bibitem{schopmeyer}
\BIBentryALTinterwordspacing
S.~A. Schopmeyer and D.~Lirman, ``Occupation dynamics and impacts of damselfish
  territoriality on recovering populations of the threatened staghorn coral,
  acropora cervicornis,'' Nov 215AD. [Online]. Available:
  \url{https://journals.plos.org/plosone/article?id=10.1371/journal.pone.0141302}
\BIBentrySTDinterwordspacing

\bibitem{corpuz}
F.~J. {Corpuz}, P.~{Naval}, E.~{Capili}, J.~{Jauod}, R.~J. {Judilla}, and
  M.~{Soriano}, ``Coral reef mosaicking using teardrop and fast image
  labeling,'' in \emph{2012 Oceans - Yeosu}, 2012, pp. 1--6.

\bibitem{dana}
D.~Manogan, P.~Cabaitan, K.~Stiefel, R.~R. Pineda, and P.~J. Naval, ``Tracking
  damselfish shoals.'' 2020, unpublished.

\bibitem{carpenter}
K.~Carpenter and V.~Springer, ``The center of the center of marine shore fish
  biodiversity: The philippine islands,'' \emph{Environmental Biology of
  Fishes}, vol.~72, pp. 467--480, 04 2005.

\bibitem{lam}
E.~Lam, ``Combining gray world and retinex theory for automatic white balance
  in digital photography,'' 07 2005, pp. 134 -- 139.

\bibitem{fischler}
\BIBentryALTinterwordspacing
M.~A. Fischler and R.~C. Bolles, ``Random sample consensus: A paradigm for
  model fitting with applications to image analysis and automated
  cartography,'' \emph{Commun. ACM}, vol.~24, no.~6, p. 381–395, Jun. 1981.
  [Online]. Available: \url{https://doi.org/10.1145/358669.358692}
\BIBentrySTDinterwordspacing

\bibitem{derpanis_2010}
\BIBentryALTinterwordspacing
K.~G. Derpanis, ``Overview of the ransac algorithm,'' May 2010. [Online].
  Available: \url{http://www.cse.yorku.ca/~kosta/CompVis\_Notes/ransac.pdf}
\BIBentrySTDinterwordspacing

\bibitem{rieucau}
G.~Rieucau, J.~Kiszka, J.~Castillo, J.~Mourier, K.~Boswell, and M.~Heithaus,
  ``Using unmanned aerial vehicle (uav) surveys and image analysis in the study
  of large surface-associated marine species: a case study on reef sharks
  carcharhinus melanopterus shoaling behaviour,'' \emph{Journal of Fish
  Biology}, vol.~93, 09 2018.

\end{thebibliography}
\end{document}